%% file: main.tex
\documentclass{ws-procs11x85}
\usepackage{ws-procs-thm}           
\usepackage{wrapfig}

\begin{document}

\title{One-Versus-Others Attention: Scalable Multimodal Integration \\ for Biomedical Data}

\author{Michal Golovanevsky, Eva Schiller, Akira Nair, Eric Han, Ritambhara Singh $^\ddag$ $^\dag$*}

\address{Department of Computer Science, Brown University,\\
$^\ddag$ and Center for Computational Molecular Biology, Brown University\\
Providence, RI 02912, USA\\
$^\dag$ E-mail: ritambhara@brown.edu}



\author{Carsten Eickhoff*}

\address{School of Medicine,\\
Institute for Bioinformatics and Medical Informatics,\\
University of Tübingen,\\
Tübingen, 72074, Germany\\
E-mail: c.eickhoff@acm.org \\
*Co-corresponding authors}

\begin{abstract}
Multimodal models have become increasingly important as they surpass single-modality approaches on diverse tasks ranging from question-answering to disease diagnosis. Despite the importance of multimodal learning, existing efforts focus on vision-language applications, where the number of modalities rarely exceeds four (images, text, audio, video). However, data in healthcare domain, may include many more modalities like X-rays, PET scans, MRIs, genetic screening, genomic data, and clinical notes, creating a need for both efficient and accurate data integration. Many state-of-the-art multimodal models rely on cross-attention or self-attention for effective data integration, which do not scale well for applications with more than two modalities. The complexity per layer of computing attention in either paradigm is, at best, quadratic with respect to the number of modalities, posing a computational bottleneck that impedes broad adoption. To address this, we propose a new attention mechanism, One-Versus-Others (OvO) attention, that scales \textit{linearly} with the number of modalities, thus offering a significant reduction in computational complexity compared to existing multimodal attention methods. Using three clinical datasets with multiple diverse modalities, we show that our method decreases computation costs while maintaining or increasing performance compared to popular integration techniques. Across all clinical datasets, OvO reduced the number of required floating point operations (FLOPs) by at least 91.98\%, demonstrating its significant impact on efficiency and enabling multi-modal predictions in healthcare.\footnote{\label{note1}Code and Appendix are available at \url{https://github.com/rsinghlab/OvO}}
\end{abstract}

\keywords{Multimodal learning; deep learning; attention mechanism; clinical decision support.}

\copyrightinfo{\copyright\ 2024 The Authors. Open Access chapter published by World Scientific Publishing Company and distributed under the terms of the Creative Commons Attribution Non-Commercial (CC BY-NC) 4.0 License.}

\section{Introduction}
\begin{figure*}[!ht]
\centering
\includegraphics[width=\textwidth]{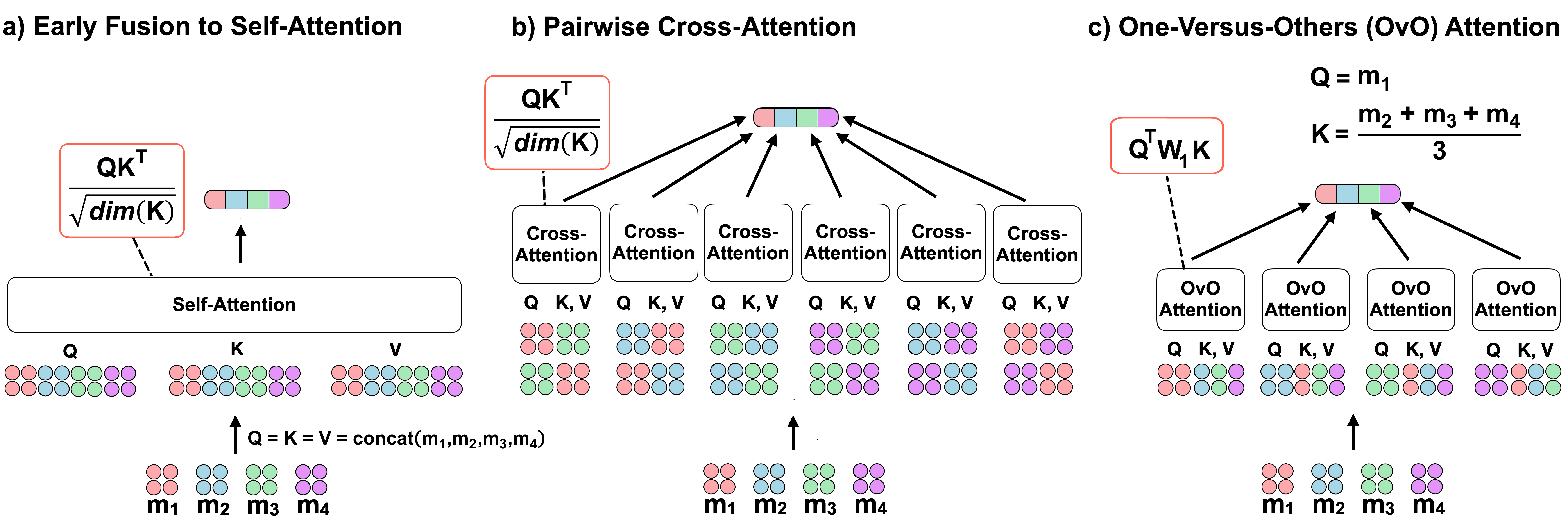}\label{fig:model_overview}
\caption{\textbf{Integration scheme comparison}. (a) Early fusion to self-attention with scaled dot product attention \cite{vaswani2017attention}, and (b) Pairwise cross-attention integration with scaled dot product attention \cite{vaswani2017attention}. (c) Our proposed method, One-Versus-Others (OvO), does not rely on pairwise interactions or long concatenated sequences but rather captures all modalities in a single attention score. A modality embedding is represented by $m_i$ and $W$ is a learnable parameter (see Section \ref{sec:ovo}).}
\end{figure*}

Multimodal learning has emerged as a promising approach, which enables joint learning from multiple modalities of data (e.g., text and images). Combining different modalities allows for a more comprehensive and accurate understanding of tasks such as clinical decision support \cite{ming2022deep, golovanevsky2022multimodal, hayat2022medfuse}, image and video captioning \cite{yu2019multimodal, seo2022end}, audio-visual speech recognition \cite{sterpu2018attention}, and sentiment analysis \cite{poria2018multimodal}. Multimodal learning has been explored through various methods in machine learning and deep learning. While feature-level integration was mostly used in more traditional machine learning algorithms, Neural Networks have allowed for the intermediate fusion of modalities at any layer and late fusion at the decision-making stage. However, these fusion paradigms lack a key component - capturing explicit interaction between modalities. For example, in detecting Alzheimer's Disease, genetic features help reinforce and ground the clinical information and thus lead to more robust decision-making \cite{golovanevsky2022multimodal}. Such relevant interactions can be captured through the attention mechanism. Popular multimodal models, such as LXMERT \cite{tan2019lxmert} and BLIP \cite{li2022blip}, use a fusion method that captures interactions between modalities using \textit{cross-attention}. On the other hand, models such as VisualBERT \cite{li2019visualbert} and LLaVA \cite{liu2024visual} use early fusion, where vision and language inputs are concatenated early to learn multimodal through \textit{self-attention}. The clinical domain embraced these approaches, with multimodal models like MedFuseNet \cite{sharma2021medfusenet} and ARMOUR \cite{liu2023attention} using cross-attention for medical vision question answering and mortality prediction. In parallel, models such as BioViL-T \cite{bannur2023learning} and MMBERT \cite{khare2021mmbert} employ early fusion through self-attention for disease prediction and report generation.

However, both self-attention and cross-attention grow quadratically in computational burden with the number of modalities, posing a scalability challenge. While in popular vision-language integration tasks, the number of modalities rarely exceeds four (images, text, audio, video), a significant bottleneck can arise in other domains. The healthcare domain exemplifies this issue, as a single task may involve integrating data from complex and rich sources spanning multiple modalities from radiology, pathology, genomics, genetics, and clinical data. Therefore, with the influx of many modalities, the use of cross-attention or self-attention will remain limited in the clinical domain as their computational demands escalate even further.
To address this gap, we propose a new attention mechanism, One-Versus-Others (OvO) attention. OvO attention is calculated by comparing \textit{one} modality against a combined representation of all \textit{other} modalities (hence the name, One-Versus-Others). Our approach significantly reduces computational complexity as it grows linearly with the number of modalities (see Section \ref{sub:complex}). Figure \ref{fig:model_overview} sketches a four-modality example to demonstrate the difference between our approach (scales linearly) and self-attention/cross-attention (scales quadratically). OvO is a general attention scheme that can be integrated into existing clinical multimodal architectures instead of cross-attention or self-attention. 

We first, present a complexity analysis and validate it through a simulated dataset. Our simulation results show scalability gains in an extreme multimodal setting (20 modalities). Next, we use three diverse clinical datasets that vary in modalities, encoder types (pre-trained and not), number of samples, and tasks to show our model's improved scalability in different clinical multimodal settings. Our results demonstrate that our method dramatically decreases computation costs (offering at minimum a 91.98\% reduction in computations), compared to self-attention and cross-attention while maintaining or even exceeding performance. 

Overall, OvO is a novel attention scheme for multimodal integration that scales linearly to the number of modalities, enabling the practical application of deep learning models in healthcare, where computational efficiency and accuracy are vital for deployment. 

\section{Related work}

Multimodal attention-based models are increasingly pivotal in clinical decision support systems, paralleling their widespread use in vision-language applications. In the medical domain, these models have shown remarkable utility in diverse scenarios, such as cancer classification \cite{li2021robust}, biomarker discovery \cite{braman2021deep, ilyin2004biomarker}, prognosis prediction \cite{schulz2021multimodal, silva2020pan}, and more. These applications highlight the versatility and potential of multimodal learning in handling complex and rich medical data. The attention mechanism serves as a core component in these models. Attention measures the similarity among individual representations, like word embedding vectors or, in the multimodal scenario, modality-specific embeddings. Each input embedding can assume one of three roles: (1) Query ($Q$), representing the current focus of attention when compared against other input embeddings; (2) Key ($K$), signifying an input embedding being compared to the Query; and (3) Value ($V$), which contributes to computing the output for the Query. 

Commonly, the representations from each modality in the multimodal models are passed through one of two paradigms - early fusion followed by self-attention or fusion through cross-attention. The early fusion group (e.g., Transformer-based models like UNITER \cite{chen2020uniter}, VisualBERT \cite{li2019visualbert}, LLaVA \cite{liu2024visual}, BioViL-T \cite{bannur2023learning}, MedViLL \cite{moon2022multi}, etc.) concatenates the visual embeddings and the textual embeddings as a single input, before passing the inputs through attention (see Figure \ref{fig:model_overview} (a)). Given modalities $\mathrm{m_{1}}$ and $\mathrm{m_{2}}$, queries ($Q$), keys ($K$), and values ($V$) are computed from their concatenated sequence (e.g., $Q_{1,2} = concat(m_{1}, m_{2})$). The final output from a standard Transformer block is denoted by $\mathrm{Z}$, Equation \ref{eq:single_stream_example} shows the early fusion paradigm. 

\begin{equation}
\label{eq:single_stream_example}
\scalebox{1.0}{$\left\{
\begin{aligned}
\begin{array}{l}
\mathrm{Z_{1,2}} = \mathit{Multiheaded\ Attention}\left(\mathrm{Q}_{\mathrm{1,2}}, \mathrm{K}_{\mathrm{1,2}}, \mathrm{V}_{\mathrm{1,2}}\right) \\
\mathrm{Z} = \mathit{Transformer}(\mathrm{Z_{1,2}})
\end{array}
\end{aligned}
\right.$}
\end{equation}

The cross-attention scheme (used in Transformer-based models like ViLBERT \cite{lu2019vilbert}, LXMERT \cite{tan2019lxmert}, MedFuseNet \cite{sharma2021medfusenet}, MADDi \cite{golovanevsky2022multimodal}, etc.) inputs each modality into its own Transformer, the outputs of which are fed to a cross-modal Transformer (see Figure \ref{fig:model_overview} (b)). For such models, the cross-modal interactions are captured in a pairwise manner through cross-attention, where queries ($Q$), keys ($K$), and values ($V$) are computed from the modality inputs ($\mathrm{m_{1}}$ and $\mathrm{m_{2}}$), and then the keys and values from each modality are fed to the multi-headed attention block of the other modality. The output, $Z$, is shown in Equation \ref{eq:dual_stream_example}.
\begin{equation}
\label{eq:dual_stream_example}
\left\{\begin{array}{l}
\mathrm{Z}_{\mathrm{1}} = \mathit{Multiheaded\ Attention}\left(\mathrm{Q}_{\mathrm{2}}, \mathrm{K}_{\mathrm{1}}, \mathrm{V}_{\mathrm{1}}\right) \\
\mathrm{Z}_{\mathrm{2}} = \mathit{Multiheaded\ Attention} \left(\mathrm{Q}_{\mathrm{1}}, \mathrm{K}_{\mathrm{2}}, \mathrm{V}_{\mathrm{2}}\right) \\
\mathrm{Z} = \mathit{Transformer}\left(\mathit{concat}\left(\mathrm{Z}_{\mathrm{1}},\mathrm{Z}_{\mathrm{2}}\right)\right)
\end{array}\right.
\end{equation}

While the early fusion and cross-attention paradigms could be extended to three modalities, seen in TriBERT \cite{rahman2021tribert} and VATT \cite{akbari2021vatt}, these models face scalability challenges for more than three modalities. Cross-attention methods can leverage joint representations formed from cross-attention but do not scale well to larger numbers of modalities as they are computed in a pairwise fashion. Thus, if there are $k$ modalities, computing pairwise fusion between each pair will result in $k \choose 2$ matrix computations. Moreover, attention is not a symmetric calculation, which means that most commonly, it is computed bi-directionally (e.g., image to text and text to image), leading to an even greater computational burden. Early fusion involves the concatenation of modalities before the Transformer layer, which similarly does not scale well with the number of modalities. Self-attention is quadratic with respect to sequence length \cite{vaswani2017attention}, and since early fusion methods concatenate inputs before attention, the computational complexity will increase quadratically as the number of modalities increases (see Section \ref{sub:complex}). Furthermore, concatenation is not order invariant, making the ordering of modalities an important consideration, potentially requiring similar bi-directional computations as cross-attention. Our integration method, OvO, addresses the limitations mentioned above in a scalable and domain-agnostic manner. 

\section{Methods}

\subsection{One-Versus-Others (OvO) attention}\label{sec:ovo}
We propose a new attention mechanism, One-Versus-Others (OvO) Attention, which grows linearly with the number of modalities rather than quadratically, as would be the case for cross-attention or self-attention (see Section \ref{sub:complex}). OvO computes attention between one modality at a time with respect to all other modalities. Given modality $m_i$, which is an embedding obtained from a dedicated encoder (e.g., CNN, ClinicalBERT, etc.) and $i \in {1, 2, \dots, k}$ where $k$ is the number of modalities, OvO takes in one modality and computes the dot product against all the other modalities with a weight matrix $W_i$. $W_i$ is a learnable parameter that can help scale the importance of each attention calculation (see Figure \ref{fig:model_overview} (c)) and can learn interactions throughout training. The modality embeddings $m_i$ in OvO attention function like queries, while the weighted sum of the other modalities behaves like keys and values, akin to the dot-product attention mechanisms, with the naming choice of $m_i$ adapted for multimodal applications. This dot product The similarity score function, representing the degree of alignment between the chosen modality and others, calculated for modality $m_i$ with respect to a set of other modalities ($m_j: j \neq i$) is shown in Equation~\ref{eq:score}. This produces a vector of scores which measure the relevance of $m_i$ with respect to the other modalities. The context vector in OvO for modality $m_i$, which is a combined representation of information from the other modalities, is shown in Equation~\ref{eq:ovo}:

\begin{equation}
\label{eq:score}
\mathit{score}\left(m_i,\{m_j:\ j\ \neq\ i\}\right)=m_i^T\ W_i\ \frac{\sum_{j\ \neq\ i}^{k}\ \ m_j}{k-1}
\end{equation}

\begin{equation}
    \label{eq:ovo}
     OvO\left(m_i,\{m_j:\ j\ \neq\ i\}\right) =
     \mathit{softmax}(\mathit{score}\left(m_i,\{m_j:\ j\ \neq\ i\}\right)) \cdot m_i
\end{equation}

In Equation~\ref{eq:ovo}, the softmax is applied across the input dimension of the embeddings, normalizing across the attention scores. The result is then multiplied by the original modality embeddings to compute the final output. We chose to sum over the ``other'' modalities instead of concatenation because: (1) the concatenation vector will continue to increase in length with the number of modalities, which will result in a less scalable framework; (2) concatenation is not invariant to the order of modalities, which could affect the model prediction, whereas a sum provides position invariance. 
 
Unlike cross-attention and self-attention, OvO provides a more interpretable mechanism for analyzing interactions between multiple modalities. In cross-attention, interactions are captured in isolated pairs (e.g., $m_1$ compared to $m_2$ or $m_1$ compared to $m_3$), limiting the ability to see how one modality integrates information from all others. Similarly, self-attention condenses modality interactions into a single operation, which can obscure explicit cross-modal interactions and make it difficult to disentangle their contributions. OvO, however, creates one attention matrix per modality, with each modality interacting with all others through the learnable weight matrix $W_{i}$, which ensures a flexible and adaptive scaling mechanism. In future work, $W_{i}$ will be studied to better understand the relative importance of each modality's contributions to the final prediction.


\subsection{Multi-headed OvO Attention}
We extend OvO attention to the multi-headed attention framework to directly compare with early fusion through self-attention and pairwise cross-attention. Multi-headed attention allows the model to attend to the input embeddings in different ways simultaneously. This is achieved by splitting the input embeddings into multiple linear projections, each processed independently through a self-attention mechanism. The outputs of each attention head are then combined to obtain the final output of the multi-headed attention layer. Formally, taking the input modality $m_i$ with respect to a set of other modalities ($m_j: j \neq i$), the multi-headed attention layer for OvO attention is defined as follows:

\begin{equation}
\left\{\begin{array}{l}
{\mathit{MultiheadedOvO}}(m_i, \{m_j:\ j\ \neq\ i\}) \\
= \mathit{concat}(\mathit{head}_1, \ldots, \mathit{head}_h)W^O \\
\mathit{head}_k = OvO(m_iW_k^{m_i}, \{m_jW_k^{m_j}:\ j\ \neq\ i\})
\end{array}\right.
\end{equation}

Here, $h$ is the number of attention heads, $W_k$ is a learnable weight matrix for the $k$-th attention head, $W^O$ is a learnable weight matrix that projects the concatenated outputs of the attention heads back to the original dimension, and Equation \ref{eq:ovo} defines OvO Attention.

\subsection{Model Complexity}\label{sub:complex}
This section highlights the complexities of the two commonly used paradigms: early fusion followed by self-attention and pairwise cross-attention, as well as our One-Versus-Others (OvO) attention. Table \ref{tab:complexity_comparison} summarizes the complexity per layer. Let \( k \) represent the number of modalities, \( n \) be the feature-length of each modality (assuming equal), and \( d \) be the representation dimension of the respective weight matrices. As established in \cite{vaswani2017attention}, self-attention has complexity of $\mathcal{O}(n^2 \cdot d)$. In the multimodal case, self-attention concatenates modalities before attention, leading to a sequence length of \( k \cdot n \), influencing the quadratic term. Thus, the complexity of self-attention is $\mathcal{O}((k \cdot n)^2 \cdot d) = \mathcal{O}(k^2 \cdot n^2 \cdot d)$. Cross-attention computes attention over all pairwise permutations of modalities: \( _{k}P_{2}  = \frac{k!}{(k-2)!} = k (k - 1)\). Thus, the number of operations required by  cross-attention is $\mathcal{O}(k \cdot (k-1) \cdot n^2 \cdot d) = \mathcal{O}((k^2 - k) \cdot n^2 \cdot d)$. When focusing on the fastest-growing terms in big \(O\) notation, the final complexity per layer simplifies to $\mathcal{O}(k^2 \cdot n^2 \cdot d)$. One-Versus-Others (OvO) Attention requires one attention calculation per modality, making it linear with respect to \( k \). Thus, the complexity per layer for OvO is $ \mathcal{O}(k \cdot n^2 \cdot d)$. Appendix Section 1 provides step-by-step details for the complexity calculations.

\begin{table}[h]
\tbl{\textbf{Per-Layer complexities of model paradigm.} Let \( k \) be the number of modalities, \( n \) the feature-length of a modality, and \( d \) the representation dimension.}
{\begin{tabular}{@{}lcccr@{}}\toprule
Model & Complexity Per Layer \\ \colrule
Self-Attention & \( \mathcal{O}(k^2 \cdot n^2 \cdot d) \) \\
Cross-Attention & \( \mathcal{O}(k^2 \cdot n^2 \cdot d) \) \\
One-Versus-Others (OvO) Attention & \( \mathcal{O}(\textbf{k} \cdot n^2 \cdot d) \) \\ \botrule
\end{tabular}}
\label{tab:complexity_comparison}
\end{table}

\subsection{Illustration through simulation}
To illustrate the linearity of OvO compared to the other integration paradigms, we simulated 20 artificial modalities. We consider two classes: (1) 20 random feature values that sum up to 1.0, and, (2) 20 random feature values that are each less than 0.15. These classes were created such that the correct label can only be inferred after inspecting all features. For example, 0.14 is less than 0.15, but it could also be a value that adds to 1. For more details on the simulation dataset and how the threshold was chosen, see Appendix Section 2.

\begin{figure}[h]
\centerline{
\includegraphics[width=0.48\textwidth]{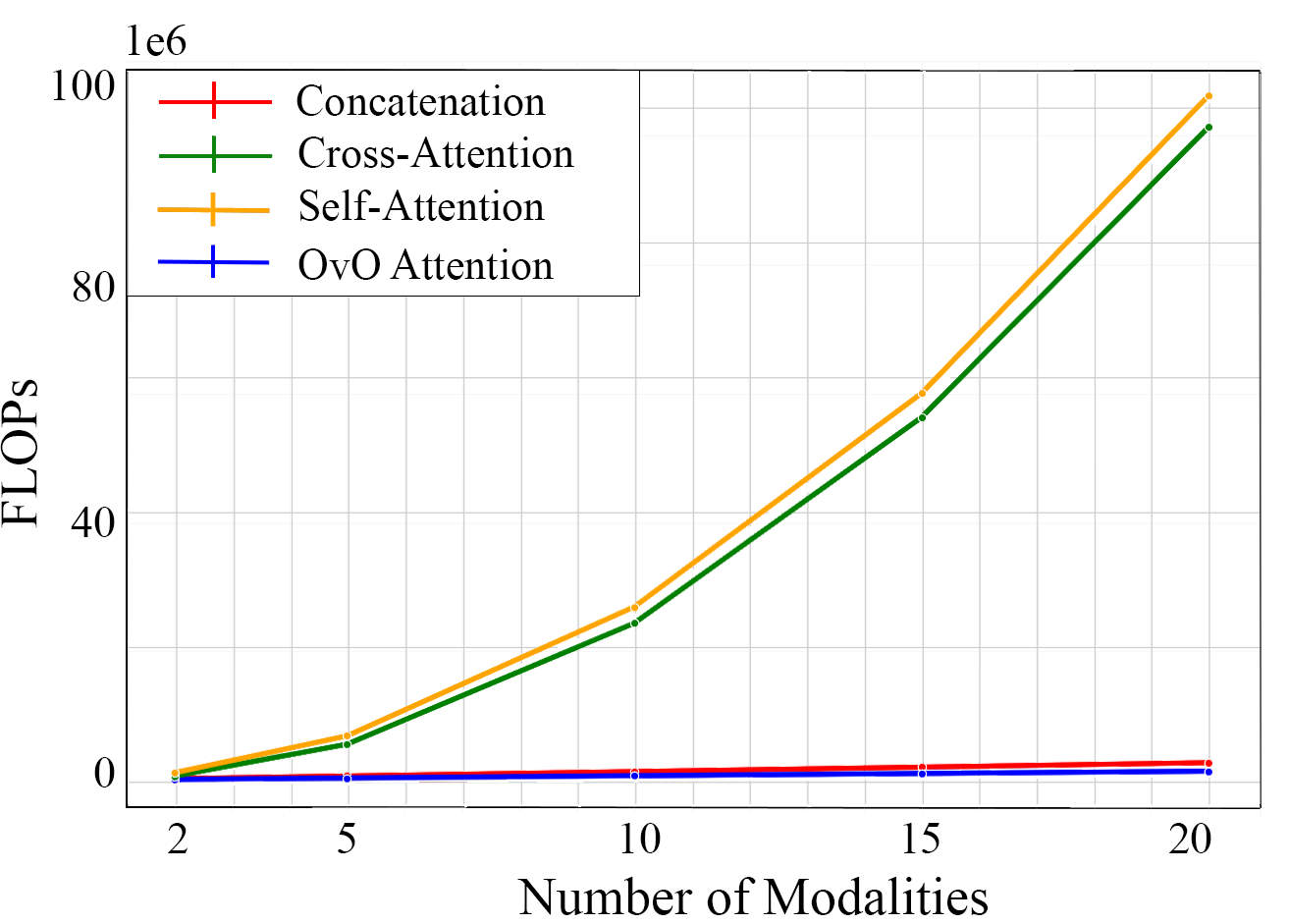}
}
\caption{\textbf{The impact of using OvO attention to fuse simulated data.} Using FLOPs as a measure of compute, we demonstrate that OvO grows linearly with respect to the number of modalities, while self and cross-attention grow quadratically.}
\label{simulation_flops}
\end{figure}

Each value was then vectorized by sampling randomly around the selected number, such that each modality is a vector of size 20 rather than a single number, leading to a combined total of 400 features. Overall, the dataset contains 2,000 samples (1,000 for each class). Our constructed simulation dataset tests the scaling capabilities of our method to an extent that real-world datasets do not usually reach. 

We examine the computation cost across the three integration methods using 2, 5, 10, 15, and 20 simulated modalities. Most notably, while self-attention and cross-attention grow quadratically with respect to the number of modalities, $k$, (\( \mathcal{O}(k^2 \cdot n^2 \cdot d) \)), our method scales linearly (\( \mathcal{O}(k \cdot n^2 \cdot d) \)), as shown in Figure~\ref{simulation_flops}. 


\section{Experiments}

We used three diverse clinical datasets to examine our method against three standard integration techniques: concatenation with no attention (baseline), early fusion with self-attention, and pairwise cross-attention. These clinical tasks feature a range of rich modalities that, despite their high integration costs, remain essential to solve. 


\subsection{Dataset descriptions}

\subsubsection{MIMIC-IV and CXR data}
MIMIC-IV \cite{johnson2023mimic} covers 431K visits for 180K patients admitted to the ICU in the Beth Israel Deaconess Medical Center. MIMIC Chest X-ray (MIMIC-CXR) \cite{johnson2019mimic} contains chest radiographs in DICOM format with free-text radiology reports. The dataset contains 377,110 images corresponding to 227,835 radiographic studies performed at Beth Israel Medical Center. We follow the pre-processing of MedFuse \cite{mohammed2021stacking} to extract the clinical time-series data from MIMIC-IV along with the associated chest X-ray images in MIMIC-CXR. We further expand the number of modalities by adding a demographics table and discharge notes, resulting in four modalities. We also follow MedFuse in the construction of the phenotyping task. The goal of this multi-label classification task is to predict whether a set of 25 chronic, mixed, and acute care conditions are assigned to a
patient in a given ICU stay. This is a 25-class multi-label task with four modalities. 

\subsubsection{The Alzheimer's Disease Prediction Of Longitudinal Evolution (TADPOLE) data}
The Alzheimer’s Disease Neuroimaging Initiative (ADNI) \cite{petersen2010alzheimer} database provides neuroimaging data, cognitive test scores, biomarker profiles, and genetic information for Alzheimer’s disease (AD), mild cognitive impairment (MCI), and normal patients. We use the processed data from the Alzheimer’s Disease Prediction Of Longitudinal Evolution (TADPOLE) challenge \cite{marinescu2019tadpole}. We focus on a one-time diagnosis prediction task, using the most recent available data point for each patient across all modalities. This ensures that each data sample includes information from the same time point, which aligns with our goal of evaluating cross-modal integration rather than longitudinal progression. We utilize six modalities that have the least missing information per patient: cognitive tests - neuropsychological tests administered by a clinical expert; MRI ROIs (generated from Freesurfer) - measures of brain structural integrity; FDG PET ROI averages - measure cell metabolism, where cells affected by AD show reduced metabolism; AV45 PET ROI averages - measures amyloid-beta load in the brain; demographic information (e.g., age, gender, education); and CSF biomarkers - amyloid and tau levels in the cerebrospinal fluid. The preprocessing provided by TADPOLE turned every modality into a tabular form (including imaging). After removing patients with missing modalities, we had 767 MCI patients, 493 normal patients, and 143 AD patients. Thus, this is a three-class classification task with six modalities. 

\subsubsection{eICU data}
The eICU collaborative database includes data from ICUs across the continental United States between 2014 and 2015 \cite{pollard2018eicu}. It consists of tables linked through a patient unit stay ID. For our mortality prediction task, we focus on six tabular modalities: patient, diagnosis, treatment, medication, lab, and apacheApsVar tables. The apacheApsVar table contains numerical variables used to calculate the Acute Physiology Score (APS), an established method within the Acute Physiology Age Chronic Health Evaluation (APACHE) system for summarizing patient's severity of illness on ICU admission and predicting outcomes. The patient table includes demographic, admission, and discharge details, and is used to determine mortality status. The diagnosis table lists active diagnoses for each patient, the treatment table includes active treatments, and the medication table contains active medication orders. We extract the features from these tables by one-hot encoding the relevant conditions, treatment types, and drug names, respectively. The lab table includes lab results, with features extracted by summing commonly recorded lab types. Our dataset includes 75,845 unique patients with 93,784 ICU stays, 86,012 recorded as alive and 7,772 as dead. This is a 2-class classification task with six modalities.

\subsection{Baselines}
Our multimodal baselines include a conventional concatenation fusion with no attention, early fusion followed by self-attention, and pairwise cross-attention fusion. The architectures of all models are identical except for their integration stage. For example, since modality-specific encoders can produce different dimension sizes, we add a linear layer before integration to create the same input dimensions. Although this step is not strictly necessary for concatenation, we still add the layer there so that no additional factors influence computation costs and performance. While there are many multimodal Transformers available for the vision-language domain, our focus is on examining the underlying fusion mechanism and creating a general integration paradigm for any application, especially ones outside of vision-language. In Appendix Section 5, we touch on the limitations of our experiments and future work that we did not cover. 

\subsection{Implementation details}\label{implementation}
For the MIMIC dataset, we follow the established train, validation, and test split in Hayat \textit{et al} \cite{hayat2022medfuse}. Similarly, for the TADPOLE task, we use the provided data splits but add a constraint that repeating patients cannot appear across data splits to avoid information leakage.
In the other datasets, for consistency, we randomly sampled 80\% of the data for the training set and 10\% each for test and validation sets, as there was not an established split. 
To evaluate our model against other integration techniques, we use the domain-accepted metrics for each task: For MIMIC and eICU, we use area under the receiver operating characteristic (AUROC) and area under Precision-Recall (PR) curve (AUPRC) as established in past works \cite{hayat2022medfuse, sheikhalishahi2020benchmarking}; For TADPOLE we use the multi-class area under the receiver operating curve (mAUC) and the overall balanced classification accuracy (BCA), as established by the competition creators \cite{marinescu2019tadpole}. For all datasets, we used the number of floating-point operations (FLOPs) as the measure of runtime complexity. FLOPs were measured per sample and reported as the difference between concatenation, the simplest integration setting, and multimodal attention ($\Delta$FLOPs). 

\subsection{Hyperparameter Tuning}\label{hyper}
Our hyperparameter tuning scheme was consistent for each dataset and each model. For each experiment, we used the evaluation metrics on the validation set to determine the best hyperparameters. We tuned the learning rate (0.01 - 1 x $10^{-8}$, dividing by 10 for each interval), batch size (16, 32, 64, 128), epochs (200 epochs with early stopping if validation performance did not increase for 5 epochs), and number of attention heads for OvO, self-attention, and cross-attention models (1, 2, 4, 8, 16). For the neural network encoders, we tuned the number of linear layers ranging from 1 to 4. Similarly, for the convolutional neural network, we tuned the number of convolution layers ranging from 1 to 4. For compute times and GPU details used for hyperparameter tuning, see Appendix Section 2. Lastly, we randomly picked 10 random seeds for every experiment - once the best hyperparameters were picked, ten models initialized with those seeds and parameters were run. Then, using the trained ten models, we evaluated on the test set and took the average of the 10 runs along with the standard deviation, which is reported in Section \ref{sec:results}.

\section{Results}\label{sec:results}
Using three real-world clinical datasets, diverse in terms of the number of modalities, feature space, and classification tasks, we demonstrate that our method consistently and drastically reduces computational costs compared to early fusion and pairwise fusion while simultaneously maintaining or enhancing performance. This is demonstrated on one four-modality dataset and two six-modality datasets.

For the four-modality MIMIC task, we used pre-trained ClinicalBERT model for the text modality and fine-tuned it for the unimodal baseline and the multimodal task, separately, ensuring adaptation in each setting. For all other modalities, we used the appropriate neural network architecture (i.e., CNN for images, LSTM for time series, and a multi-layer perception for all tabular data). We perform significance testing between OvO attention and the next best-performing model, detailed in Appendix Section 3. 

\begin{table}[h]
\tbl{\textbf{MIMIC IV+CXR results}.(*) FLOPs were measured per sample and reported as the difference between concatenation and multimodal attention. We offer improved performance across all metrics and reduce FLOPs by at least 93.73\% compared to self and cross-attention.}
{\begin{tabular}{@{}llcccr@{}}\toprule
Model & Modalities & $\downarrow$ $\Delta$ FLOPs & $\uparrow$ AUROC & $\uparrow$ AUPRC\\ \colrule
LSTM & Time Series & - & 58.8 \textpm 0.6 & 28.5 \textpm 0.4\\
CNN & Images & - & 56.9 \textpm 0.3 & 26.7 \textpm 0.2\\
Neural Net & Demographics & - & 64.1 \textpm 0.4 & 32.4 \textpm 0.3\\
ClinicalBERT & Text & - & 79.3 \textpm 0.4 & 58.7 \textpm 0.3\\ \colrule
Concatenation & All & * & 82.7 \textpm 0.6 & 65.1 \textpm 1.8\\
Cross-Attention & All & 52,723,712 & 78.2 \textpm 2.1 & 54.1 \textpm 2.7\\
Self-Attention & All & 67,633,152 & 78.5 \textpm 2.0 & 55.7 \textpm 3.1\\ \colrule
\textbf{OvO Attention} & \textbf{All} & \textbf{4,227,072} & \textbf{83.6 \textpm 1.1} & \textbf{66.2 \textpm 2.6}\\ \botrule
\end{tabular}}\label{mimic-table}
\end{table}

\begin{table}[h]
\tbl{\textbf{TADPOLE results}. (*) FLOPs were measured per sample and reported as the difference between concatenation and multimodal attention. We offer improved performance across all metrics and reduce FLOPs by at least 95.45\% compared to self and cross-attention.}
{\begin{tabular}{@{}llcccr@{}}\toprule
Model & Modalities & $\downarrow$ $\Delta$ FLOPs & $\uparrow$ MAUC & $\uparrow$ BCA\\ \colrule
Neural Net & AV45 PET ROI & - & 63.5 \textpm 3.1 & 56.4 \textpm 3.8\\
Neural Net & CSF Biomarkers & - & 64.4 \textpm 1.1 & 53.6 \textpm 2.7\\
Neural Net & MRI ROIs & - & 67.0 \textpm 1.3 & 57.2 \textpm 1.0\\
Neural Net & FDG PET ROI & - & 66.6 \textpm 0.3 & 60.8 \textpm 0.7\\
Neural Net & Demographics & - & 74.6 \textpm 0.9 & 62.0 \textpm 0.6\\
Neural Net & Cognitive Tests & - & 97.8 \textpm 0.2 & 88.6 \textpm 0.7\\ \colrule
Concatenation & All & * & 97.7 \textpm 0.8 & 91.9 \textpm 1.9\\
Cross-Attention & All & 8,921,088 & 97.1 \textpm 0.6 & 90.7 \textpm 1.7\\
Self-Attention & All & 9,633,792 & 94.8 \textpm 1.1 & 86.6 \textpm 2.6\\ \colrule
\textbf{OvO Attention} & \textbf{All} & \textbf{405,504} & \textbf{98.3 \textpm 0.4} & \textbf{93.0 \textpm 1.4}\\ \botrule
\end{tabular}}\label{tadpole-table}
\end{table}

The results on MIMIC are presented in Table~\ref{mimic-table}, clearly demonstrating the scalability and performance advantages of OvO attention. OvO's 4,227,072 FLOPs notably reduce computational costs compared to cross-attention (52,723,712 FLOPs) and self-attention (67,633,152 FLOPs), achieving reductions by \textbf{91.98\%} and \textbf{93.75\%}, respectively, thus highlighting OvO’s superior efficiency. The unimodal results show that the textual modality is most valuable in phenotype prediction, and ClinicalBERT alone performs better than self-attention and cross-attention. This indicates that the added complexity and forced interactions are not necessarily conducive to result quality. However, OvO attention can extract information from the other modalities for a significant performance increase rather than a decrease (p-value \textless 0.01, see Appendix Section 3). 

For the six-modality Alzheimer's detection task from TADPOLE, we show our results in Table \ref{tadpole-table}. OvO's 405,504 FLOPs significantly undercut cross-attention (8,921,088 FLOPs) and self-attention (9,633,792 FLOPs), achieving reductions of \textbf{95.45\%} and \textbf{95.79\%}, respectively, highlighting OvO's remarkable efficiency.
Similarly to the MIMIC results, the unimodal results show that the cognitive tests modality is most valuable in disease prediction, and performs on its own better than self-attention and cross-attention. However, OvO attention can extract information from the other modalities for a significant performance increase rather than a decrease (p-value \textless 0.01).

Lastly, the results on the six-modality eICU mortality prediction task are shown in Table \ref{eicu-table}, demonstrating the scalability and performance advantages of OvO attention. 

\begin{table}[h]
\tbl{\textbf{eICU results}. We report the average of 10 random seeds for AUROC and AUPRC, along with standard deviations. (*) FLOPs were measured per sample and reported as the difference between concatenation and multimodal attention. We offer improved performance across all metrics and reduce FLOPs by at least 95.12\% compared to self and cross-attention.}
{\begin{tabular}{@{}llcccr@{}}\toprule
Model & Modalities & $\downarrow$ $\Delta$ FLOPs & $\uparrow$ AUROC & $\uparrow$ AUPRC\\ \colrule
Neural Net & Demographics & - & 50.2 \textpm 0.6 & 91.8 \textpm 0.2\\
Neural Net & Medication & - & 56.3 \textpm 1.3 & 93.1 \textpm 0.3\\
Neural Net & Diagnosis & - & 58.2 \textpm 2.1 & 93.3 \textpm 0.4\\
Neural Net & Treatment & - & 66.1 \textpm 0.5 & 94.8 \textpm 0.1\\
Neural Net & APACHE APS & - & 77.6 \textpm 0.2 & 97.0 \textpm 0.1\\
Neural Net & Laboratory & - & 81.5 \textpm 0.4 & 97.0 \textpm 0.1\\ \colrule
Concatenation & All & * & 81.7 \textpm 1.6 & 97.5 \textpm 0.3\\
Cross-Attention & All & 129,957,888 & 77.6 \textpm 1.6 & 95.4 \textpm 0.3\\
Self-Attention & All & 151,781,376 & 80.2 \textpm 2.0 & 96.8 \textpm 0.4\\ \colrule
\textbf{OvO Attention} & \textbf{All} & \textbf{6,340,608} & \textbf{82.5 \textpm 0.9} & \textbf{97.8 \textpm 0.2}\\ \botrule
\end{tabular}}\label{eicu-table}
\end{table}
OvO's 6,340,608 FLOPs significantly undercut those of cross-attention (129,957,888 FLOPs) and self-attention (151,781,376 FLOPs), achieving reductions of approximately \textbf{95.12\%} and \textbf{95.82\%}, respectively, thereby highlighting OvO's efficiency.
Mirroring the trends observed in the MIMIC and TADPOLE datasets, we note a dominant unimodal modality, specifically Lab modality, in this experiment as well. While concatenating modalities does enhance performance, this improvement is not seen in self and cross-attention models. In contrast, OvO attention not only reflects these performance gains but does so significantly (p-value \textless 0.01). We hypothesize that this is due to the overfitting of more complex integration frameworks of self and cross-attention on relatively smaller clinical datasets. OvO, in its simplicity akin to concatenation, manages to strike a balance by maintaining flexibility and capturing inter-modal interactions through its attention mechanism, thus offering an edge in performance without excessive complexity.

In summary, across diverse clinical datasets and modalities, OvO attention consistently outperforms traditional fusion techniques in both predictive performance and computational efficiency, underlining its robustness in handling complex multimodal healthcare data.



\section{Conclusion}

We present One-Versus-Others (OvO), a new scalable multimodal attention mechanism. The proposed formulation significantly reduces the computational complexity compared to the widely used early fusion through self-attention and cross-attention methods. Notably, OvO achieves, at minimum, a reduction of 91.98\% in FLOPs when benchmarked against self and cross-attention methods across a range of clinical datasets containing up to six modalities. We provide both a detailed theoretical complexity analysis and empirical evidence from a simulated experiment, illustrating that OvO's computational demand scales linearly with the number of modalities, in contrast to the quadratic scaling observed in other methods. Our proposed method provides a way to overcome one of the major challenges associated with multimodal datasets - computational resource demand and cost, thus enabling adoption in resource-constrained domains, such as clinical decision support. 
Overall, the results unequivocally establish that OvO not only significantly reduces computational expenses but also exceeds the performance of existing state-of-the-art fusion methodologies.

\newpage
\input{appendix}

\bibliographystyle{plain}
\bibliography{bib}

\end{document}

%% file: appendix.tex










\appendix{}

\section{Computational Complexity Analysis for Multimodal Integration Schemes}
\label{complex_appendix}

In this section, we present the step-by-step details of the computational complexity analysis presented in Section 3.3. The analysis is done with respect to the size of the input modalities associated with the three paradigms used in our experimental setting: early fusion followed by self-attention, cross-modal attention, and One-Versus-Others (OvO) Attention. 

\subsection{Early Fusion}

The early fusion approach involves first combining the modalities and then processing the concatenated sequence with the self-attention mechanism.

\textbf{Step 1: Concatenation of Modalities.}

Let $k$ be the number of modalities and $n$ be the feature-length of each modality. 
\[ \mbox{Total length after concatenation = } k \times n \]
The complexity for this operation is linear:
\[ \mathcal{O}(k \cdot n) \]

\textbf{Step 2: Compute Queries, Keys, and Values.}

The self-attention mechanism derives queries (Q), keys (K), and values (V) for the concatenated sequence (length $k \cdot n$) using linear transformations with representation dimension, $d$. The complexity of each transformation operation is:
\[ \mathcal{O}(k \cdot n \cdot d) \]

\textbf{Step 3: Compute Attention Scores.}

Attention scores are computed by taking the dot product of queries and keys. The self-attention mechanism has quadratic complexity with respect to the sequence length and linear complexity with respect to the representation dimension $d$ \cite{vaswani2017attention}. Thus, given the concatenated sequence's length of \( k \cdot n \) and the dimension of the keys and queries \( d \), the complexity of this step is:
\[ \mathcal{O}((k \cdot n)^2 \cdot d) = \mathcal{O}(k^2 \cdot n^2 \cdot d) \]

\textbf{Step 4: Calculate the Weighted Sum for Outputs.}

For each of the \( k \cdot n \) positions in the concatenated sequence, we compute the softmax of the attention scores to produce the attention weights. These weights are then multiplied with their corresponding \( d \)-dimensional values to compute the weighted sum, which becomes the output. The computational complexity of these operations is:
\[ \mathcal{O}(k^2 \cdot n^2 \cdot d) \]

When combining all steps, the dominating terms in the computational complexity stem from the attention scores' computation and the weighted sum, culminating in an overall complexity of:
\[ \mathcal{O}(k^2 \cdot n^2 \cdot d) \]

\subsection{Cross-modal Attention}

For cross-modal attention, each modality attends to every other modality. 

\textbf{Step 1: Compute  Queries, Keys, and Values for Inter-Modal Attention.}

From a given modality, compute a query (Q), and from the remaining \( k-1 \) modalities, compute keys (K) and values (V). Keys, queries, and values are obtained using linear transformations with representation dimension $d$. The complexity of each transformation operation is: 

\[ \mathcal{O}(n \cdot d) \mbox{ for each query, key, value set} \]
Considering all modalities:
\[ \mathcal{O}(k \cdot (k-1) \cdot n \cdot d) \]

The term \(k \cdot (k-1)\) comes from the number of pairwise permutations of \( k \), given by \( _{k}P_{2}  = \frac{k!}{(k-2)!} = k (k - 1)\).

\textbf{Step 2: Calculate Attention Scores for Inter-Modal Attention.}\\
The queries and keys from different modalities are used to compute attention scores, which represent how much one modality should attend to another. 
\[ \mathcal{O}(n^2 \cdot d) \mbox{ for each pair of modalities \cite{vaswani2017attention} } \]
Considering all modalities:
\[ \mathcal{O}(k \cdot (k-1) \cdot n^2 \cdot d) \]

\textbf{Step 3: Calculate the Weighted Sum for Outputs.}\\
For every modality interaction, calculate the softmax of the attention scores to obtain the attention weights. These weights are then used in conjunction with the values vector to derive the weighted sum for the output:

\[ \mathcal{O}(n^2 \cdot d) \mbox{ for each pair of modalities} \]
Considering all modalities:
\[ \mathcal{O}(k \cdot (k-1) \cdot n^2 \cdot d) \]

When evaluating all steps together, the dominating factors in computational complexity arise from the computation of attention scores and the weighted sum. Thus, the collective complexity for cross-modal attention, where each modality attends to every other, equates to:
\[ \mathcal{O}(k \cdot (k-1) \cdot n^2 \cdot d) = \mathcal{O}((k^2 - k) \cdot n^2 \cdot d) \]

For the complexity of cross-modal attention, the dominant term is \(k^2\). The \(k-1\) term effectively becomes a constant factor in relation to \(k^2\). As \(k\) tends toward larger values, the difference between \(k^2\) and \(k^2 - k\) diminishes. This is a consequence of the principles of big \(O\) notation, which focuses on the fastest-growing term in the equation while dismissing constant factors and lower-order terms. As a result, for asymptotic analysis, the complexity 
\[ \mathcal{O}(k^2 - k) \cdot n^2 \cdot d \] 
can be simplified to: 
\[ \mathcal{O}(k^2 \cdot n^2 \cdot d) \].

\subsection{One-Versus-Others (OvO) Attention Complexity}

\textbf{Step 1: Averaging of "Other" Modalities.}\\
Let $k$ be the number of modalities and $n$ be the feature-length of each modality. For each modality \( m_i \), averaging over the other \( k-1 \) modalities results in a complexity of:
\[ \mathcal{O}(n) \]
Given that this needs to be computed for all \( k \) modalities:
\[ \mathcal{O}(k \cdot n) \]

\textbf{Step 2: Calculate Attention Scores with Shared Weight Matrix W.}\\
The modality vector \( m_i \) and the average of "other" modalities, $\frac{\sum_{j\ \neq\ i}^{n}\ \ m_j}{n-1}$, are used to compute attention scores, which represent how much one modality should attend to the others. Multiplication with the weight matrix \( W \) (with representation dimension $d$) and the dot product with the summed modalities lead to:

\[ \mathcal{O}(n^2 \cdot d) \]
Considering this operation for all $k$ modalities:
\[ \mathcal{O}(k \cdot n^2 \cdot d) \]

\textbf{Step 3: Calculate the Weighted Sum for Outputs.}\\
For every modality interaction, calculate the softmax of the attention scores to obtain the attention weights. These weights are then used in conjunction with the \( m_i \) vector (analogous the values (V) vector) to derive the weighted sum for the output:

\[ \mathcal{O}(n^2 \cdot d) \mbox{ for each pair of modalities} \]
Considering all modalities:
\[ \mathcal{O}(k \cdot n^2 \cdot d) \]

When evaluating all steps together, the dominating factors in computational complexity arise from the computation of attention scores. Thus, the collective complexity for cross-modal attention, where each modality attends to every other, equates to:
\[ \mathcal{O}(k \cdot n^2 \cdot d) \]

In summary, One-Versus-Others (OvO) Attention exhibits a computational complexity that grows linearly with respect to the number of modalities (\( \mathcal{O}(k \cdot n^2 \cdot d) \)). In contrast, both early fusion through self-attention and cross-attention approaches demonstrate quadratic growth with respect to the number of modalities (\( \mathcal{O}(k^2 \cdot n^2 \cdot d) \)). This makes OvO a more scalable option for multimodal integration.

\section{Simulation Dataset Details}\label{app:simulation}
We consider two classes: (1) 20 random feature values that sum up to 1.0, and, (2) 20 random feature values
that are each less than 0.15. The threshold was chosen at 0.15 because if 0.10 was the threshold, the mean of the 20 values would be 0.05, and thus, the sum would also be very close to 1, on average. This would render the task too difficult, and there would not be a significant difference between the samples across the two labels. Setting the threshold to 0.2 would render the task too easy, as on average, the numbers are consistently greater in the second class and the classes could be differentiated using only one modality. Thus, we chose 0.15 as the threshold.

\section{Compute Resources}\label{appendix:compute} 
For each experiment, we use one NVIDIA GeForce RTX 3090 GPU. For the MIMIC task, single-modality models ran for roughly 40 minutes, and multi-modal models ran for roughly 55 minutes on average. For the eICU, the single modality pre-trained models ran for roughly 50 minutes, the single modality neural network ran for a minute, and the multi-modal models ran for approximately an hour on average. For the TADPOLE task, single-modality models ran for 5 minutes, while multi-modal models ran for roughly 15 minutes on average. In the simulation dataset, the maximum modalities was 20 which took our model, OvO, roughly 2 minutes to run, while the cross-modal attention baseline took about 20 minutes to run on average.

\begin{table}[h]
\tbl{Average runtimes for different tasks and model types using one NVIDIA GeForce RTX 3090 GPU.}
{\begin{tabular}{@{}lcccr@{}}\toprule
\textbf{Task} & \textbf{Models} & \textbf{Runtime (minutes)}\\ \colrule
MIMIC & Unimodal & 40 \\
      & Multimodal & 55 \\
eICU  & Unimodal pre-trained & 50 \\
      & Unimodal neural net & 1 \\
      & Multimodal & 60 \\
TADPOLE & Unimodal & 5 \\
        & Multimodal & 15 \\
Simulation & OvO (20 modalities) & 2 \\
           & Cross and Self-attention  & 20 \\ \botrule
\end{tabular}}\label{tab:runtime}
\end{table}

\section{Significance Testing}\label{appendix:stat}
We use a t-test to determine if there is a significant difference the performance metrics (AUROC, AUPRC, MAUC, BCA) means between OvO attention and the next best-performing multimodal model. Our sample size is 10 from each group, as we initialized the models with 10 random seeds. For the MIMIC IV and CXR dataset, we compare against self-attention as it performed the second best after OvO. Using an $\alpha$ = 0.01, we have evidence to reject the null hypothesis and conclude that there is a statistically significant difference in means between single-attention and OvO attention. The p-value for the AUROC scores is 0.00363 and the p-value for AUPRC is 0.000948.  
For the TADPOLE challenge, we compare against cross-attention as it performed the second best after OvO. We get a p-value for MAUC scores of $2.09e^{-7}$ and a p-value of $8.19e^{-12}$ for BCA. Thus, we demonstrate a statistically significant difference in MAUC and BCA means between self-attention and OvO attention. 
Lastly, for the eICU dataset, we compare against cross-attention as it performed the second best after OvO. We get a p-value for AUROC scores of $2.00e^{-15}$ and a p-value of $8.24e^{-14}$ for AUPRC. Thus, we demonstrate a statistically significant difference in AUROC and AUPRC means between self-attention and OvO attention.

\begin{table}[h]
\tbl{Results of t-tests comparing the performance metrics between OvO attention and the next best-performing multimodal models across different datasets.}
{\begin{tabular}{@{}llcccr@{}}\toprule
\textbf{Dataset} & \textbf{Comparison Model} & \textbf{Metric} & \textbf{p-value} \\ \colrule
MIMIC & Self-attention & AUROC & 0.00363 \\
      &                & AUPRC & 0.000948 \\
TADPOLE & Cross-attention & MAUC  & $2 \times 10^{-7}$ \\
        &                 & BCA   & $8 \times 10^{-12}$ \\
eICU    & Cross-attention & AUROC & $2 \times 10^{-15}$ \\
        &                 & AUPRC & $8 \times 10^{-14}$ \\ \botrule
\end{tabular}}\label{tab:t-test-results}
\end{table}

\section{Limitations}
This paper’s primary goal is to address one of the major challenges associated with multimodal datasets - computational resource demand and cost. While we demonstrate the scalability of the OvO attention mechanism and its efficiency in handling multiple modalities, we did not conduct experiments focused on interpretability, which is also a crucial aspect of multimodal learning in the biomedical space. The potential for OvO to reveal modality importance through the learninable $W_i$ parameter is promising, but further experiments are required to explore this interpretability in a meaningful way. Additionally, while we compared OvO to popular attention mechanisms like self-attention and cross-attention, there are other attention variants that could serve as future baselines, though they are not widely used in multimodal models. Lastly, the publicly available multimodal datasets often limit the diversity of modalities we can test, with many datasets primarily consisting of tabular data. 

\newpage



%% file: main.bbl
\begin{thebibliography}{10}

\bibitem{akbari2021vatt}
Hassan Akbari, Liangzhe Yuan, Rui Qian, Wei-Hong Chuang, Shih-Fu Chang, Yin Cui, and Boqing Gong.
\newblock Vatt: Transformers for multimodal self-supervised learning from raw video, audio and text.
\newblock {\em Advances in Neural Information Processing Systems}, 34:24206--24221, 2021.

\bibitem{bannur2023learning}
Shruthi Bannur, Stephanie Hyland, Qianchu Liu, Fernando Perez-Garcia, Maximilian Ilse, Daniel~C Castro, Benedikt Boecking, Harshita Sharma, Kenza Bouzid, Anja Thieme, et~al.
\newblock Learning to exploit temporal structure for biomedical vision-language processing.
\newblock In {\em Proceedings of the IEEE/CVF Conference on Computer Vision and Pattern Recognition}, pages 15016--15027, 2023.

\bibitem{braman2021deep}
Nathaniel Braman, Jacob~WH Gordon, Emery~T Goossens, Caleb Willis, Martin~C Stumpe, and Jagadish Venkataraman.
\newblock Deep orthogonal fusion: multimodal prognostic biomarker discovery integrating radiology, pathology, genomic, and clinical data.
\newblock In {\em Medical Image Computing and Computer Assisted Intervention--MICCAI 2021: 24th International Conference, Strasbourg, France, September 27--October 1, 2021, Proceedings, Part V 24}, pages 667--677. Springer, 2021.

\bibitem{chen2020uniter}
Yen-Chun Chen, Linjie Li, Licheng Yu, Ahmed El~Kholy, Faisal Ahmed, Zhe Gan, Yu~Cheng, and Jingjing Liu.
\newblock Uniter: Universal image-text representation learning.
\newblock In {\em European conference on computer vision}, pages 104--120. Springer, 2020.

\bibitem{golovanevsky2022multimodal}
Michal Golovanevsky, Carsten Eickhoff, and Ritambhara Singh.
\newblock Multimodal attention-based deep learning for alzheimer’s disease diagnosis.
\newblock {\em Journal of the American Medical Informatics Association}, 29(12):2014--2022, 2022.

\bibitem{hayat2022medfuse}
Nasir Hayat, Krzysztof~J Geras, and Farah~E Shamout.
\newblock {MedFuse: Multi-modal fusion with clinical time-series data and chest X-ray images}.
\newblock In {\em Machine Learning for Healthcare Conference}, pages 479--503. PMLR, 2022.

\bibitem{ilyin2004biomarker}
Sergey~E Ilyin, Stanley~M Belkowski, and Carlos~R Plata-Salam{\'a}n.
\newblock Biomarker discovery and validation: technologies and integrative approaches.
\newblock {\em Trends in biotechnology}, 22(8):411--416, 2004.

\bibitem{johnson2019mimic}
Alistair Johnson, Matt Lungren, Yifan Peng, Zhiyong Lu, Roger Mark, Seth Berkowitz, and Steven Horng.
\newblock Mimic-cxr-jpg-chest radiographs with structured labels.
\newblock {\em PhysioNet}, 2019.

\bibitem{johnson2023mimic}
Alistair~EW Johnson, Lucas Bulgarelli, Lu~Shen, Alvin Gayles, Ayad Shammout, Steven Horng, Tom~J Pollard, Sicheng Hao, Benjamin Moody, Brian Gow, et~al.
\newblock Mimic-iv, a freely accessible electronic health record dataset.
\newblock {\em Scientific data}, 10(1):1, 2023.

\bibitem{khare2021mmbert}
Yash Khare, Viraj Bagal, Minesh Mathew, Adithi Devi, U~Deva Priyakumar, and CV~Jawahar.
\newblock Mmbert: Multimodal bert pretraining for improved medical vqa.
\newblock In {\em 2021 IEEE 18th International Symposium on Biomedical Imaging (ISBI)}, pages 1033--1036. IEEE, 2021.

\bibitem{li2021robust}
BOCHONG LI, Toshiya Nakaguchi, Yuichiro Yoshimura, and Ping Xuan.
\newblock Robust multi-modal prostate cancer classification via feature disentanglement and dual attention.
\newblock {\em Transactions of Japanese Society for Medical and Biological Engineering}, (Abstract):308--308, 2021.

\bibitem{li2022blip}
Junnan Li, Dongxu Li, Caiming Xiong, and Steven Hoi.
\newblock Blip: Bootstrapping language-image pre-training for unified vision-language understanding and generation.
\newblock In {\em International conference on machine learning}, pages 12888--12900. PMLR, 2022.

\bibitem{li2019visualbert}
Liunian~Harold Li, Mark Yatskar, Da~Yin, Cho-Jui Hsieh, and Kai-Wei Chang.
\newblock Visualbert: A simple and performant baseline for vision and language.
\newblock {\em arXiv preprint arXiv:1908.03557}, 2019.

\bibitem{liu2024visual}
Haotian Liu, Chunyuan Li, Qingyang Wu, and Yong~Jae Lee.
\newblock Visual instruction tuning.
\newblock {\em Advances in neural information processing systems}, 36, 2024.

\bibitem{liu2023attention}
Jinghui Liu, Daniel Capurro, Anthony Nguyen, and Karin Verspoor.
\newblock Attention-based multimodal fusion with contrast for robust clinical prediction in the face of missing modalities.
\newblock {\em Journal of Biomedical Informatics}, 145:104466, 2023.

\bibitem{lu2019vilbert}
Jiasen Lu, Dhruv Batra, Devi Parikh, and Stefan Lee.
\newblock Vilbert: Pretraining task-agnostic visiolinguistic representations for vision-and-language tasks.
\newblock {\em Advances in neural information processing systems}, 32, 2019.

\bibitem{marinescu2019tadpole}
R{\u{a}}zvan~V Marinescu, Neil~P Oxtoby, Alexandra~L Young, Esther~E Bron, Arthur~W Toga, Michael~W Weiner, Frederik Barkhof, Nick~C Fox, Polina Golland, Stefan Klein, et~al.
\newblock Tadpole challenge: Accurate alzheimer’s disease prediction through crowdsourced forecasting of future data.
\newblock In {\em Predictive Intelligence in Medicine: Second International Workshop, PRIME 2019, Held in Conjunction with MICCAI 2019, Shenzhen, China, October 13, 2019, Proceedings 2}, pages 1--10. Springer, 2019.

\bibitem{ming2022deep}
Yue Ming, Xiying Dong, Jihuai Zhao, Zefu Chen, Hao Wang, and Nan Wu.
\newblock Deep learning-based multimodal image analysis for cervical cancer detection.
\newblock {\em Methods}, 205:46--52, 2022.

\bibitem{mohammed2021stacking}
Mohanad Mohammed, Henry Mwambi, Innocent~B Mboya, Murtada~K Elbashir, and Bernard Omolo.
\newblock A stacking ensemble deep learning approach to cancer type classification based on tcga data.
\newblock {\em Scientific reports}, 11(1):1--22, 2021.

\bibitem{moon2022multi}
Jong~Hak Moon, Hyungyung Lee, Woncheol Shin, Young-Hak Kim, and Edward Choi.
\newblock Multi-modal understanding and generation for medical images and text via vision-language pre-training.
\newblock {\em IEEE Journal of Biomedical and Health Informatics}, 26(12):6070--6080, 2022.

\bibitem{petersen2010alzheimer}
Ronald~Carl Petersen, Paul~S Aisen, Laurel~A Beckett, Michael~C Donohue, Anthony~Collins Gamst, Danielle~J Harvey, Clifford~R Jack, William~J Jagust, Leslie~M Shaw, Arthur~W Toga, et~al.
\newblock Alzheimer's disease neuroimaging initiative (adni): clinical characterization.
\newblock {\em Neurology}, 74(3):201--209, 2010.

\bibitem{pollard2018eicu}
Tom~J Pollard, Alistair~EW Johnson, Jesse~D Raffa, Leo~A Celi, Roger~G Mark, and Omar Badawi.
\newblock The eicu collaborative research database, a freely available multi-center database for critical care research.
\newblock {\em Scientific data}, 5(1):1--13, 2018.

\bibitem{poria2018multimodal}
Soujanya Poria, Navonil Majumder, Devamanyu Hazarika, Erik Cambria, Alexander Gelbukh, and Amir Hussain.
\newblock Multimodal sentiment analysis: Addressing key issues and setting up the baselines.
\newblock {\em IEEE Intelligent Systems}, 33(6):17--25, 2018.

\bibitem{rahman2021tribert}
Tanzila Rahman, Mengyu Yang, and Leonid Sigal.
\newblock Tribert: Full-body human-centric audio-visual representation learning for visual sound separation.
\newblock {\em arXiv preprint arXiv:2110.13412}, 2021.

\bibitem{schulz2021multimodal}
Stefan Schulz, Ann-Christin Woerl, Florian Jungmann, Christina Glasner, Philipp Stenzel, Stephanie Strobl, Aur{\'e}lie Fernandez, Daniel-Christoph Wagner, Axel Haferkamp, Peter Mildenberger, et~al.
\newblock Multimodal deep learning for prognosis prediction in renal cancer.
\newblock {\em Frontiers in oncology}, 11:788740, 2021.

\bibitem{seo2022end}
Paul~Hongsuck Seo, Arsha Nagrani, Anurag Arnab, and Cordelia Schmid.
\newblock End-to-end generative pretraining for multimodal video captioning.
\newblock In {\em Proceedings of the IEEE/CVF Conference on Computer Vision and Pattern Recognition}, pages 17959--17968, 2022.

\bibitem{sharma2021medfusenet}
Dhruv Sharma, Sanjay Purushotham, and Chandan~K Reddy.
\newblock Medfusenet: An attention-based multimodal deep learning model for visual question answering in the medical domain.
\newblock {\em Scientific Reports}, 11(1):19826, 2021.

\bibitem{sheikhalishahi2020benchmarking}
Seyedmostafa Sheikhalishahi, Vevake Balaraman, and Venet Osmani.
\newblock Benchmarking machine learning models on multi-centre eicu critical care dataset.
\newblock {\em Plos one}, 15(7):e0235424, 2020.

\bibitem{silva2020pan}
Lu{\'\i}s A~Vale Silva and Karl Rohr.
\newblock Pan-cancer prognosis prediction using multimodal deep learning.
\newblock In {\em 2020 IEEE 17th International Symposium on Biomedical Imaging (ISBI)}, pages 568--571. IEEE, 2020.

\bibitem{sterpu2018attention}
George Sterpu, Christian Saam, and Naomi Harte.
\newblock Attention-based audio-visual fusion for robust automatic speech recognition.
\newblock In {\em Proceedings of the 20th ACM International Conference on Multimodal Interaction}, pages 111--115, 2018.

\bibitem{tan2019lxmert}
Hao Tan and Mohit Bansal.
\newblock Lxmert: Learning cross-modality encoder representations from transformers.
\newblock {\em arXiv preprint arXiv:1908.07490}, 2019.

\bibitem{vaswani2017attention}
Ashish Vaswani, Noam Shazeer, Niki Parmar, Jakob Uszkoreit, Llion Jones, Aidan~N Gomez, {\L}ukasz Kaiser, and Illia Polosukhin.
\newblock Attention is all you need.
\newblock {\em Advances in neural information processing systems}, 30, 2017.

\bibitem{yu2019multimodal}
Jun Yu, Jing Li, Zhou Yu, and Qingming Huang.
\newblock Multimodal transformer with multi-view visual representation for image captioning.
\newblock {\em IEEE transactions on circuits and systems for video technology}, 30(12):4467--4480, 2019.

\end{thebibliography}
